\pdfoutput=1

\documentclass[11pt]{article}

\usepackage{multirow}
\usepackage{graphicx}
\usepackage{booktabs}
\usepackage{colortbl}
\usepackage{paralist}
\usepackage{enumitem}
\usepackage{amssymb}
\usepackage{hyperref}
\usepackage{bm}
\usepackage{makecell}
\usepackage{arydshln}
\usepackage{pifont}
\usepackage{longtable}
\usepackage{ACL2023}
\usepackage{times}
\usepackage{latexsym}
\usepackage[T1]{fontenc}
\usepackage[utf8]{inputenc}
\usepackage{microtype}
\usepackage{inconsolata}

\title{\textit{Bad Actor, Good Advisor:} \\Exploring the Role of Large Language Models in Fake News Detection}

\newcommand{\emailsI}{\href{mailto:hubeizhe21s@ict.ac.cn}{hubeizhe21s},\href{mailto:shengqiang18z@ict.ac.cn}{shengqiang18z},\href{mailto:caojuan@ict.ac.cn}{caojuan},\href{mailto:shiyuhui22s@ict.ac.cn}{shiyuhui22s}}

\newcommand{\emailsII}{\href{mailto:liyang23s@ict.ac.cn}{liyang23s},\href{mailto:wangdanding@ict.ac.cn}{wangdanding}}

\author{
Beizhe Hu$^{1,2}$\quad
Qiang Sheng$^{1,}$\thanks{$^*$Corresponding author.}\quad
Juan Cao$^{1,2}$\quad
Yuhui Shi$^{1,2}$ \\
{\bf Yang Li}$^{1,2}$ \quad
{\bf Danding Wang}$^{1}$ \quad
{\bf Peng Qi}$^{3}$ \\
	$^{1}$Key Lab of Intelligent Information Processing of Chinese Academy of Sciences, \\
	Institute of Computing Technology, Chinese Academy of Sciences\\
	$^{2}$University of Chinese Academy of Sciences \quad 
    $^{3}$National University of Singapore\\
	\texttt{\{\emailsI\}@ict.ac.cn}\\
	\texttt{\{\emailsII\}@ict.ac.cn}, \texttt{\href{mailto:pengqi.qp@gmail.com}{pengqi.qp}@gmail.com
 }
}

\begin{document}
\maketitle
\begin{abstract}
Detecting fake news requires both a delicate sense of diverse clues and a profound understanding of the real-world background, which remains challenging for detectors based on small language models (SLMs) due to their knowledge and capability limitations.
Recent advances in large language models (LLMs) have shown remarkable performance in various tasks, but whether and how LLMs could help with fake news detection remains underexplored.
In this paper, we investigate the potential of LLMs in fake news detection.
First, we conduct an empirical study and find that a sophisticated LLM such as GPT 3.5 could generally expose fake news and provide desirable multi-perspective rationales but still underperforms the basic SLM, fine-tuned BERT.
Our subsequent analysis attributes such a gap to the LLM's inability to select and integrate rationales properly to conclude. 
Based on these findings, we propose that current LLMs may not substitute fine-tuned SLMs in fake news detection but can be a good advisor for SLMs by providing multi-perspective instructive rationales.
To instantiate this proposal, we design an adaptive rationale guidance network for fake news detection (ARG), in which SLMs selectively acquire insights on news analysis from the LLMs' rationales.
We further derive a rationale-free version of ARG by distillation, namely ARG-D, which services cost-sensitive scenarios without querying LLMs.
Experiments on two real-world datasets demonstrate that ARG and ARG-D outperform three types of baseline methods, including SLM-based, LLM-based, and combinations of small and large language models.
\end{abstract}

\begin{figure}[t]
	\centering
	\includegraphics[width=\linewidth]{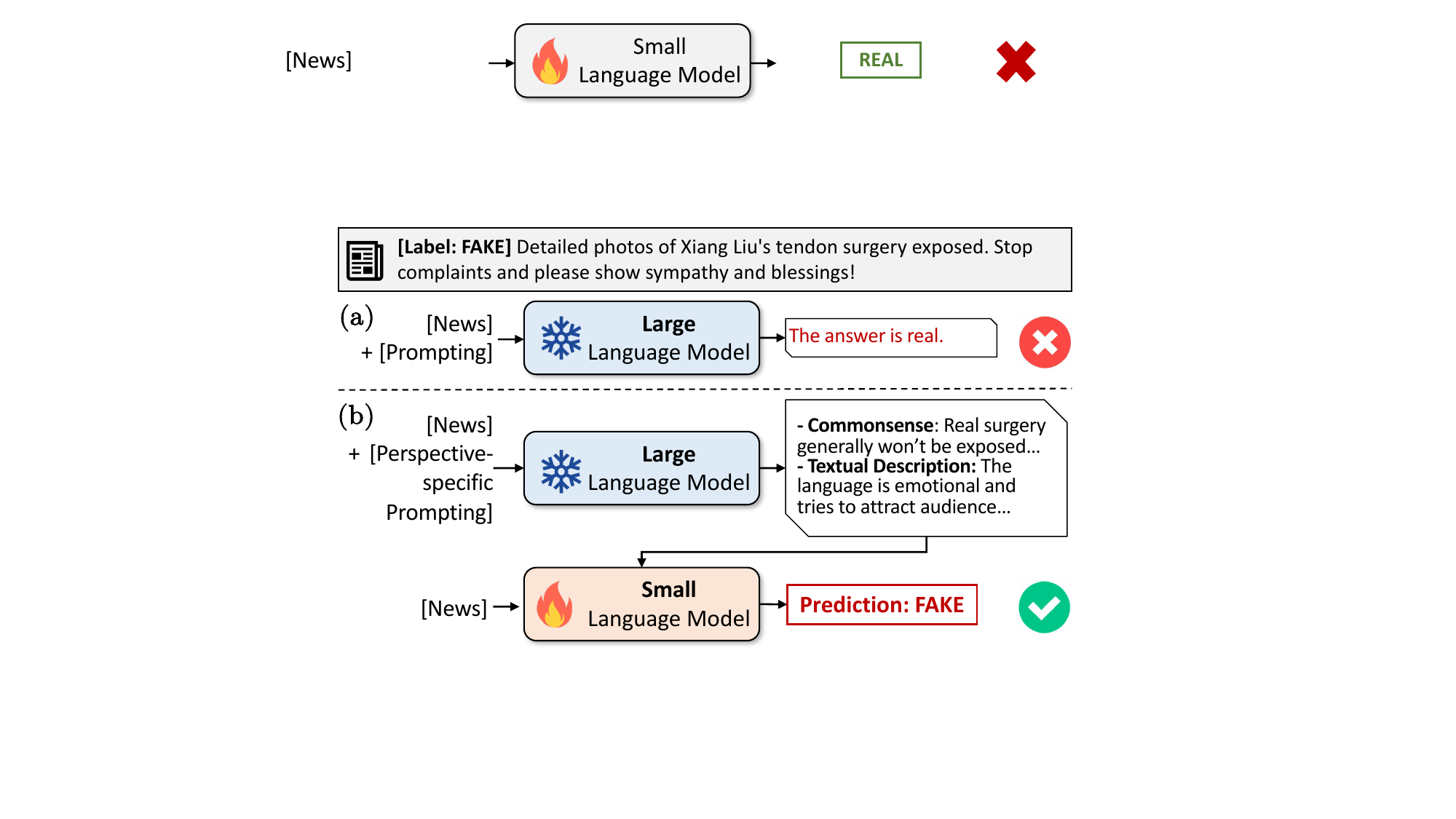}
	\caption{Illustration of the role of large language models (LLMs) in fake news detection. In this case, (a) the LLM fails to output correct judgment of news veracity but (b) helps the small language model (SLM) judge correctly by providing informative rationales.}
	\label{fig:motivating}
\end{figure}

\section{Introduction}

The wide and fast spread of fake news online has posed real-world threats in critical domains like politics~\cite{pizzagate}, economy~\cite{fake-news-economy}, and public health~\cite{infodemic}. Among the countermeasures to combat this issue, \textit{automatic fake news detection},  which aims at distinguishing inaccurate and intentionally misleading news items from others automatically, has been a promising solution in practice~\cite{shu2017survey,roth2022tweet}.

Though much progress has been made~\cite{hu2022deep}, understanding and characterizing fake news is still challenging for current models.
This is caused by the complexity of the news-faking process: Fake news creators might manipulate any part of the news, using diverse writing strategies and being driven by inscrutable underlying aims.
Therefore, to maintain both effectiveness and universality for fake news detection, 
an ideal method is required to have: 1) a delicate sense of diverse clues (\textit{e.g.}, style, facts, commonsense); and 2) a profound understanding of the real-world background. 

Recent methods~\cite{dual-emotion,fakebert,shukai-www22,ftt} generally exploit pre-trained \textbf{small language models (SLMs)}\footnote{The academia lacks a consensus regarding the size boundary between small and large language models at present, but it is widely accepted that BERT~\cite{bert} and GPT-3 family~\cite{gpt3} are respectively small and large ones~\cite{LLM-survey}.} like BERT~\cite{bert} and RoBERTa~\cite{roberta} to understand news content and provide fundamental representation, plus optional social contexts~\cite{defend,cui2022meta}, knowledge bases~\cite{declare,chef}, or news environment~\cite{newsenv} as supplements.
SLMs do bring improvements, but their knowledge and capability limitations also compromise further enhancement of fake news detectors. For example, BERT was pre-trained on text corpus like Wikipedia~\cite{bert} and thus struggled to handle news items that require knowledge not included~\cite{pref-fend}.

As a new alternative to SLMs, \textbf{large language models (LLMs)}~\cite{chatgpt,claude2,llama}, which are usually trained on the larger-scale corpus and aligned with human preferences, have shown impressive emergent abilities on various tasks~\cite{emergent-ability} and are considered promising as general task solvers~\cite{Reranker}.
However, the potential of LLMs in fake news detection remains underexplored:
\begin{compactitem}
    \item Can LLMs help detect fake news with their internal knowledge and capability?
    \item What solution should we adopt to obtain better performance using LLMs?
\end{compactitem}

To answer these two questions, we first conduct a deep investigation of the effective role of LLMs in fake news detection and attempt to provide a practical LLM-involved solution.
Unlike contemporary works~\cite{concurrent1,concurrent2} which simply prompt LLMs to provide predictions with the task instruction, we conduct a detailed empirical study to mine LLMs' potential.
Specifically, we use four typical prompting approaches (zero-shot/few-shot vanilla/chain-of-thought prompting) to ask the LLM to make veracity judgments of given news items (\figurename~\ref{fig:motivating}(a)) and find that even the best-performing LLM-based method still underperforms task-specific fine-tuned SLMs.
We then perform an analysis of the LLM-generated explanatory rationales and find that the LLM could provide reasonable and informative rationales from several perspectives.
By subsequently inducing the LLM with perspective-specific prompts and performing rule-based ensembles of judgments, we find that rationales indeed benefit fake news detection, and attribute the unsatisfying performance to the LLM's inability to select and integrate rationales properly to conclude.

Based on these findings, we propose that the current LLM may not be a good substitute for the well-fine-tuned SLM but could serve as a good advisor by providing instructive rationales, as presented in \figurename~\ref{fig:motivating}(b). 
To instantiate our proposal, we design the adaptive rationale guidance (ARG) network for fake news detection, which bridges the small and large LMs by selectively injecting new insight about news analysis from the large LM's rationales to the small LM. 
The ARG further derives the rationale-free ARG-D via distillation for cost-sensitive scenarios with no need to query LLMs.
Experiments on two real-world datasets show that ARG and ARG-D outperform existing SLM/LLM-only and combination methods.
Our contributions are as follows:

\begin{itemize}
    \item \textbf{Detailed investigation:} We investigate the effective role of LLMs in fake news detection and find the LLM is bad at veracity judgment but good at analyzing contents;
    \item \textbf{Novel and practical solution:} We design a novel ARG network and its distilled version ARG-D that complements small and large LMs by selectively acquiring insights from LLM-generated rationales for SLMs, which has shown superiority based on extensive experiments;
    \item \textbf{Useful resource:} We construct a rationale collection from GPT-3.5 for fake news detection in two languages (Chinese and English) and will make it publicly available to facilitate further research.\footnote{\url{https://github.com/ICTMCG/ARG}}
\end{itemize}

\section{Is the LLM a Good Detector?}
\label{sec:analysis}

\begin{table}[t]
\small
\centering
\setlength{\tabcolsep}{5pt}
\begin{tabular}{@{}crllllll@{}}
\toprule
\multicolumn{2}{c}{\multirow{2}{*}[-0.3em]{\textbf{\#}}} & \multicolumn{3}{c}{\textbf{Chinese}} & \multicolumn{3}{c}{\textbf{English}} \\
\cmidrule(){3-8} 
\multicolumn{2}{c}{} & \multicolumn{1}{c}{Train} & \multicolumn{1}{c}{Val} & \multicolumn{1}{c}{Test} & \multicolumn{1}{c}{Train} & \multicolumn{1}{c}{Val} & \multicolumn{1}{c}{Test} \\ \midrule
\multicolumn{2}{r}{Real} & \multicolumn{1}{r}{2,331} & \multicolumn{1}{r}{1,172} & \multicolumn{1}{r}{1,137} & \multicolumn{1}{r}{2,878} & \multicolumn{1}{r}{1,030} & \multicolumn{1}{r}{1,024} \\
\multicolumn{2}{r}{Fake} & \multicolumn{1}{r}{2,873} & \multicolumn{1}{r}{779} & \multicolumn{1}{r}{814} & \multicolumn{1}{r}{1,006} & \multicolumn{1}{r}{244} & \multicolumn{1}{r}{234} \\
\multicolumn{2}{r}{Total} & \multicolumn{1}{r}{5,204} & \multicolumn{1}{r}{1,951} & \multicolumn{1}{r}{1,951} & \multicolumn{1}{r}{3,884} & \multicolumn{1}{r}{1,274} & \multicolumn{1}{r}{1,258} \\
\bottomrule
\end{tabular}
\caption{Statistics of the fake news detection datasets.}
\label{tab:data_stats}%
\end{table}%

In this section, we evaluate the performance of the representative LLM, \textit{i.e.}, GPT-3.5 in fake news detection to reveal its judgment capability. We exploit four typical prompting approaches and perform a comparison with the SLM (here, BERT) fine-tuned on this task. Formally, given a news item $x$, the model aims to predict whether $x$ is fake or not.

\subsection{Experimental Settings}
\paragraph{Dataset}
We employ the Chinese dataset Weibo21~\cite{mdfend} and the English dataset GossipCop~\cite{fakenewsnet} for evaluation.
Following existing works~\cite{endef,mu-etal-2023-time}, we preprocess the datasets with deduplication and temporal data split to avoid possible performance overrating led by data leakage for the SLM. \tablename~\ref{tab:data_stats} presents the dataset statistics.

\paragraph{Large Language Model}
\begin{figure}[t]
	\centering
\includegraphics[width=1\linewidth]{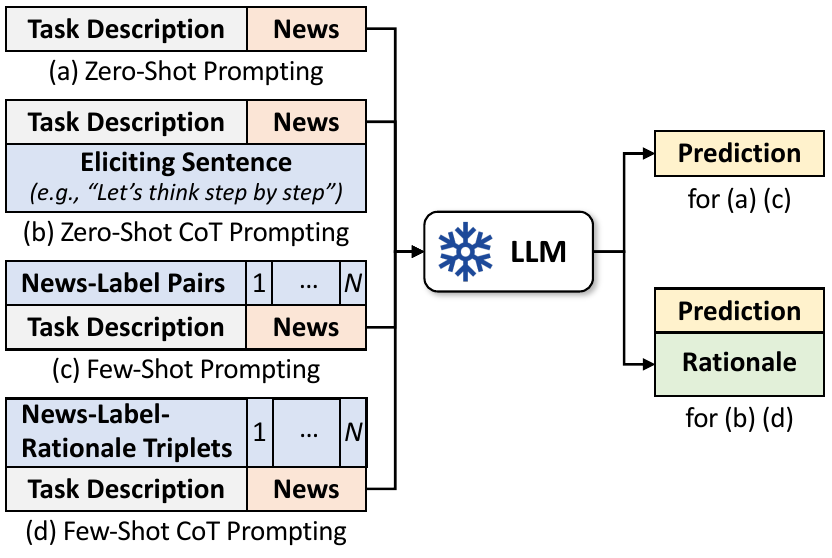}
	\caption{Illustration of prompting approaches for LLMs.}
	\label{fig:prompting}
\end{figure}
We evaluate GPT-3.5-turbo, the LLM developed by OpenAI and supporting the popular chatbot ChatGPT~\cite{chatgpt}, due to its representativeness and convenient calling. The large scale of parameters makes task-specific fine-tuning almost impossible for LLMs, so we use the prompt learning paradigm, where an LLM learns tasks given prompts containing instructions or few-shot demonstrations~\cite{prompt-survey}. In detail, we utilize the following four typical prompting approaches to elicit the potential of the LLM in fake news detection (\figurename~\ref{fig:prompting}):

\begin{itemize}
    \item \textbf{Zero-Shot Prompting}
    constructs prompt only containing the task description and the given news. To make the response more proficient and decrease the refusal ratio, we optionally adopt the role-playing technique when describing our task~\cite{role-play-1,role-play-2}.

    \item \textbf{Zero-Shot CoT Prompting}~\cite{zero-shot-cot}
    is a simple and straightforward chain-of-thought (CoT) prompting approach to encourage the LLM to reason. In addition to the elements in zero-shot prompting, it adds an eliciting sentence such as \textit{``Let's think step by step.''}

    \item \textbf{Few-Shot Prompting}~\cite{gpt3}
    provides task-specific prompts and several news-label examples as demonstrations. After preliminary tests of \{2,4,8\}-shot settings, we choose 4-shot prompting which includes two real and two fake samples.
    
    \item \textbf{Few-Shot CoT Prompting}~\cite{few-shot-cot}
    not only provides news-label examples but also demonstrates reasoning steps with previously written rationales. Here, we obtain the provided rationale demonstrations from the correct and reasonable outputs of zero-shot CoT prompting.
\end{itemize}

\paragraph{Small Language Model}
We adopt the pre-trained small language models, BERT~\cite{bert} as the representative, given its wide use in this task~\cite{fakebert,endef,newsenv}.
Specifically, we limit the maximum length of the text to 170 tokens and use \textit{chinese-bert-wwm-ext} and \textit{bert-base-uncased} from Transformers package~\cite{transformers} for the Chinese and English evaluation, respectively. We use Adam~\cite{adam} as the optimizer and do a grid search for the optimal learning rate.
We report the testing result on the best-validation checkpoint.

\begin{table}[t]
  \centering
  \small
    \begin{tabular}{cccc}
    \toprule
    \textbf{Model} & \textbf{Usage} & \textbf{Chinese} & \textbf{English} \\
    \midrule
    \multirow{4}{*}{\begin{tabular}[c]{@{}c@{}}GPT-3.5-\\turbo\end{tabular}} & Zero-Shot & 0.676  & 0.568  \\
          & Zero-Shot CoT & 0.677  & 0.666  \\
          & Few-Shot & \underline{0.725}  & 0.697  \\
          & Few-Shot CoT & 0.681     & \underline{0.702} \\
    \midrule
    \multirow{2}{*}{BERT}  & \multirow{2}{*}{Fine-tuning} & \textbf{0.753}  & \textbf{0.765}  \\
    & & \textbf{(+3.8\%)} & \textbf{(+9.0\%)} \\
    \bottomrule
    \end{tabular}
  \caption{Performance in macro F1 of the large and small LMs. The best two results are \textbf{bolded} and \underline{underlined}, respectively. The relative increases over the second-best results are shown in the brackets.}
  \vspace{-0.3cm}
  \label{tab:anal_all}%
\end{table}%

\begin{table}[t]
  \centering
  \small
  \setlength{\tabcolsep}{1.8mm}
    \begin{tabular}{@{}ccccccc@{}}
    \toprule
    \multirow{2}{*}[-0.3em]{\textbf{Perspective}} & \multicolumn{2}{c}{\textbf{Chinese}} & &  \multicolumn{2}{c}{\textbf{English}} \\
     \cmidrule{2-3}  \cmidrule{5-6} 
       & Prop. & macF1 & & Prop. & macF1 \\
       \midrule\midrule
    \textbf{Textual Description} & 65\%  & 0.706  & & 71\%  & 0.653  \\
    \midrule
    \multicolumn{6}{p{23em}}{\textbf{News:} Everyone! Don't buy cherries anymore: Cherries of this year are infested with maggots, and nearly 100\% are affected. 
    \newline{}\textbf{LLM Rationale:} ...The tone of the news is extremely urgent, seemingly trying to spread panic and anxiety.
    \newline{}\textbf{Prediction:} Fake \hspace{1em} \textbf{Ground Truth:} Fake} \\
    \midrule\midrule
    \textbf{Commonsense} & 71\%  & 0.698 &  & 60\%  & 0.680  \\
    \midrule
    \multicolumn{6}{p{23em}}{\textbf{News:} Huang, the chief of Du'an Civil Affairs Bureau, gets subsistence allowances of 509 citizens, owns nine properties, and has six wives... 
    \newline{}\textbf{LLM Rationale:} ...The news content is extremely outrageous...Such a situation is incredibly rare in reality and even could be thought impossible.
    \newline{}\textbf{Prediction:} Fake \hspace{1em} \textbf{Ground Truth:} Fake} \\
    \midrule\midrule
    \textbf{Factuality}  & 17\%  & 0.629 & & 24\%  & 0.626  \\
    \midrule
    \multicolumn{6}{p{23em}}{\textbf{News:} The 18th National Congress has approved that individuals who are at least 18 years old are now eligible to marry...
    \newline{}\textbf{LLM Rationale:} First, the claim that Chinese individuals at least 18 years old can register their marriage is real, as this is stipulated by Chinese law...
    \newline{}\textbf{Prediction:} Real \hspace{1em} \textbf{Ground Truth:} Fake} \\
    \midrule\midrule
    \textbf{Others} & 4\%   & 0.649 & & 8\%   & 0.704  \\
    \bottomrule
    \end{tabular}%
  \caption{Analysis of different perspectives of  LLM's rationales in the sample set, including the data ratio, LLM's performance, and cases. Prop.: Proportion.}
  \label{tab:anal_sampled}%
\end{table}%

\subsection{Comparison between Small and Large LMs}
\tablename~\ref{tab:anal_all} presents the performance of GPT-3.5-turbo with four prompting approaches and the fine-tuned BERT on the two datasets. We observe that:
\textbf{1) }Though the LLM is generally believed powerful, \textbf{the LLM underperforms the fine-tuned SLM using all four prompting approaches}. The SLM has a relative increase of 3.8\%$\sim$11.3\% in Chinese and 9.0\%$\sim$34.6\% in English over the LLM, indicating that the LLM lacks task-specific knowledge while the SLM learns during fine-tuning.

\textbf{2) }Few-shot versions outperform zero-shot ones, suggesting the importance of task samples. However, introducing several samples only narrow the gap with the SLM but does not lead to surpassing. 

\textbf{3) }CoT prompting brings additional performance gain in general, especially under the zero-shot setting on the English dataset (+17.3\%). However, we also observe some cases where CoT leads to a decrease. This indicates that effective use of rationales may require more careful design.

Overall, given the LLM's unsatisfying performance and higher inference costs than the SLM, the current LLM has not been a ``good enough'' detector to substitute task-specific SLMs in fake news detection.

\subsection{Analysis on the Rationales from the LLM}

\begin{table}[t]
    \centering
    \small
    \setlength{\tabcolsep}{3.5pt}
    \begin{tabular}{@{}clcc@{}}
    \toprule
    \textbf{Model} & \textbf{Usage} & \textbf{Chinese} & \textbf{English} \\
    \midrule
        \multirow{3}{*}{GPT-3.5-turbo} & Zero-Shot CoT  & 0.677 & 0.666 \\
        & \ \ from Perspective TD & 0.667 & 0.611 \\
        & \ \ from Perspective CS & 0.678 & 0.698 \\
        \midrule
        \multirow{1}{*}{BERT} & Fine-tuning  & 0.753 & 0.765 \\
        \midrule
        \multirow{2}{*}{Ensemble} & Majority Voting & 0.735 & 0.724 \\
         & Oracle Voting & 0.908 & 0.878 \\
     \bottomrule
    \end{tabular}
    \caption{Performance of the LLM using zero-shot CoT with perspective specified and other compared models. TD: Textual description; CS: Commonsense.}
    \label{tab:single-view-all}
\end{table}

Though the LLM is bad at news veracity judgment, we also notice that the rationales generated through zero-shot CoT prompting exhibit a unique multi-perspective analytical capability that is challenging and rare for SLMs.
For further exploration, we sample 500 samples from each of the two datasets and manually categorize them according to the perspectives from which the LLM performs the news analysis. Statistical results by perspectives and cases are presented in \tablename~\ref{tab:anal_sampled}.\footnote{Note that a sample may be analyzed from multiple perspectives and thus the sum of \textit{proportions} might be larger than 100\%.}
We see that:
\textbf{1)} \textbf{The LLM is capable of generating human-like rationales on news content from various perspectives}, such as textual description, commonsense, and factuality, which meets the requirement of the delicate sense of diverse clues and profound understanding of the real-world background in fake news detection.
\textbf{2)} The detection performance on the subset using certain perspectives is higher than the zero-shot CoT result on the full testing set. This indicates the potential of analysis by perspectives, though the coverage is moderate.
\textbf{3) }The analysis from the perspective of factuality leads to the performance lower than average, indicating the unreliability of using the LLM for factuality analysis based on its internal memorization. We speculate this is caused by the hallucination issue~\cite{hallucination, hallucination2}.

We further investigate the LLM's performance when asked to perform analysis from a specific perspective on the full testing set (\textit{i.e.}, 100\% coverage).\footnote{We exclude the factuality to avoid the impacts of hallucination. The eliciting sentence is ``Let's think from the perspective of [textual description/commonsense].''} From the first group in \tablename~\ref{tab:single-view-all}, we see that the LLM's judgment with single-perspective analysis elicited is still promising. Compared with the comprehensive zero-shot CoT setting, the single-perspective-based LLM performs comparatively on the Chinese dataset and is better on the English dataset (for the commonsense perspective case).
The results showcase that the internal mechanism of the LLM to integrate the rationales from diverse perspectives is ineffective for fake news detection, limiting the full use of rationales.
In this case, combining the small and large LMs to complement each other is a promising solution: The former could benefit from the analytical capability of the latter, while the latter could be enhanced by task-specific knowledge from the former.

To exhibit the advantages of this solution, we apply majority voting and oracle voting (assuming the most ideal situation where we trust the correctly judged model for each sample, if any) among the two single-perspective-based LLMs and the BERT.
Results show that we are likely to gain a performance better than any LLM-/SLM-only methods mentioned before if we could adaptively combine their advantages, \textit{i.e.}, the flexible task-specific learning of the SLM and the informative rationale generated by the LLM. That is, \textbf{the LLM could be possibly a good advisor for the SLM by providing rationales, ultimately improving the performance of fake news detection.}

\begin{figure*}[t]
	\centering
	\includegraphics[width=\textwidth]{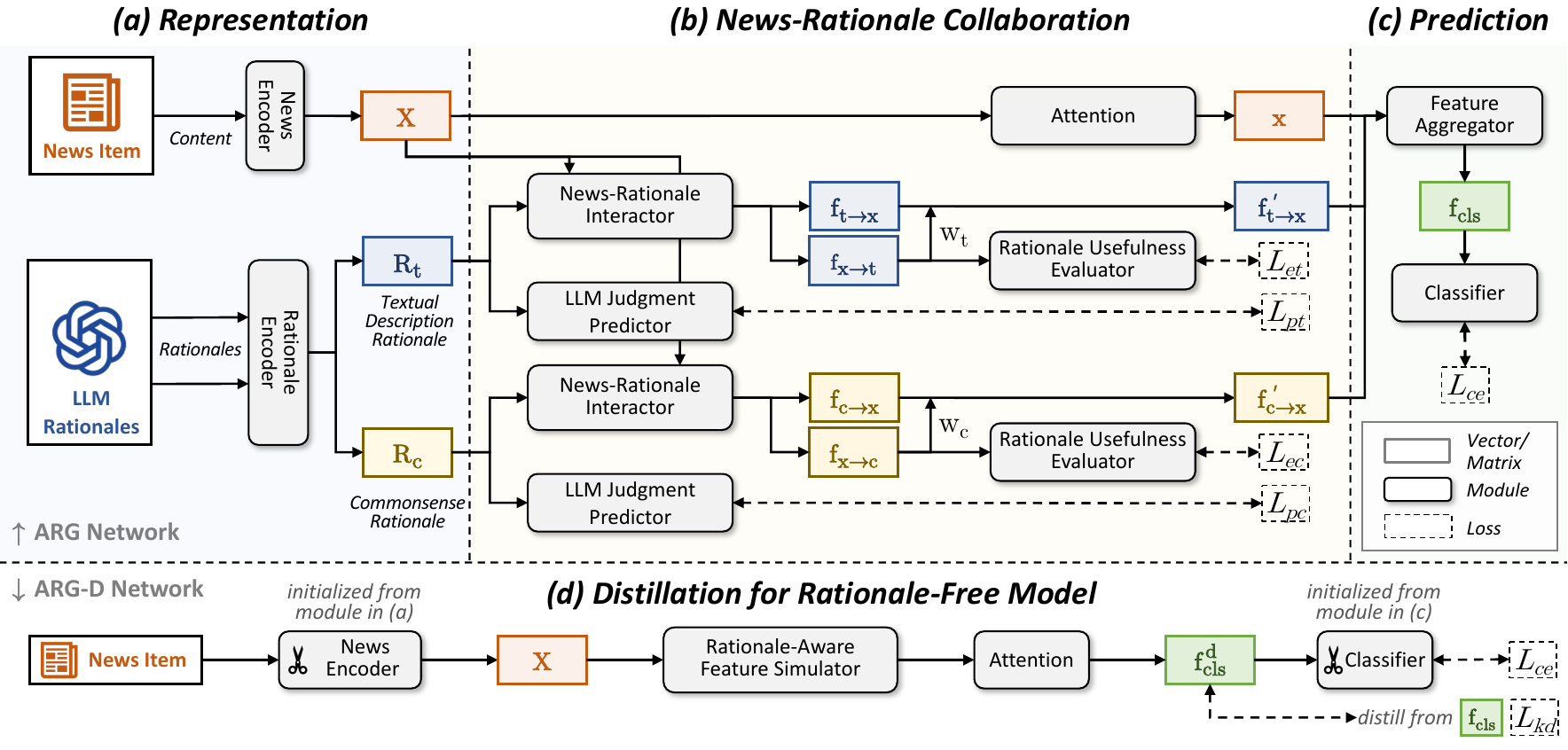}
	\caption{Overall architecture of our proposed adaptive rationale guidance (ARG) network and its rationale-free version ARG-D. In the ARG, the news item and LLM rationales are (a) respectively encoded into $\mathrm{\mathbf{X}}$ and $\mathrm{\mathbf{R_{*}}} (* \in \{t,c\})$. Then the small and large LMs collaborate with each other via news-rationale feature interaction, LLM judgment prediction, and rationale usefulness evaluation. The obtained interactive features $\mathrm{\mathbf{f^\prime_{*{\rightarrow}x}}}$ $(* \in \{t,c\})$. These features are finally aggregated with attentively pooled news feature $\mathrm{\mathbf{x}}$ for the final judgment. In the ARG-D, the news encoder and the attention module are preserved and the output of the rationale-aware feature simulator is supervised by the aggregated feature $\mathrm{\mathbf{f_{cls}}}$ for knowledge distillation.}
	\label{fig:arch}
 \vspace{-0.3cm}
\end{figure*}

\section{ARG: Adaptive Rationale Guidance Network for Fake News Detection}
Based on the above findings and discussion, we propose the adaptive rationale guidance (ARG) network for fake news detection.
\figurename~\ref{fig:arch} overviews the ARG and its rationale-free version ARG-D, for cost-sensitive scenarios.
The objective of ARG is to empower small fake news detectors with the ability to adaptively select useful rationales as references for final judgments.
Given a news item $x$ and its corresponding LLM-generated rationales $r_{t}$ (textual description) and $r_{c}$ (commonsense), the ARG encodes the inputs using the SLM at first (\figurename~\ref{fig:arch}(a)). Subsequently, it builds news-rationale collaboration via predicting the LLM's judgment through the rationale, enriching news-rationale feature interaction, and evaluating rationale usefulness (\figurename~\ref{fig:arch}(b)). The interactive features are finally aggregated with the news feature $\mathrm{\mathbf{x}}$ for the final judgment of $x$ being fake or not (\figurename~\ref{fig:arch}(c)). ARG-D is derived from the ARG via distillation for scenarios where the LLM is unavailable (\figurename~\ref{fig:arch}(d)).

\subsection{Representation}
We employ two BERT models separately as 
the news and rationale encoder to obtain semantic representations. For the given news item $x$ and two corresponding rationales $r_t$ and $r_c$, the representations are $\mathrm{\mathbf{X}}$, $\mathrm{\mathbf{R_t}}$, and $\mathrm{\mathbf{R_c}}$, respectively.

\subsection{News-Rationale Collaboration}
The step of news-rationale collaboration aims at providing a rich interaction between news and rationales and learning to adaptively select useful rationales as references, which is at the core of our design. To achieve such an aim, ARG includes three modules, as detailed 
 and exemplified using the textual description rationale branch below:

\subsubsection{News-Rationale Interaction}
To enable comprehensive information exchange between news and rationales, we introduce a news-rationale interactor with a dual cross-attention mechanism to encourage feature interactions.
The cross-attention can be described as:
\begin{equation}
\mathrm{CA(\mathrm{\mathbf{Q}},\mathrm{\mathbf{K}},\mathrm{\mathbf{V}})} = \mathrm{softmax}\left(\mathrm{\mathbf{Q^\prime}}\cdot \mathrm{\mathbf{K^\prime}}/\sqrt{d}\right) \mathrm{\mathbf{V^\prime}},
\end{equation}
where $\mathrm{\mathbf{Q^\prime}}=\mathrm{\mathbf{W}_Q}\mathrm{\mathbf{Q}}$, $\mathrm{\mathbf{K^\prime}}=\mathrm{\mathbf{W}_K}\mathrm{\mathbf{K}}$, and $\mathrm{\mathbf{V^\prime}}=\mathrm{\mathbf{W}_V}\mathrm{\mathbf{V}}$. $d$ is the dimensionality. 
Given representations of the news $\mathrm{\mathbf{X}}$ and the rationale $\mathrm{\mathbf{R_t}}$, the process is:
\begin{equation}
 \mathrm{\mathbf{f_{t{\rightarrow}x}}} = \mathrm{AvgPool}\left(\mathrm{CA(\mathrm{\mathbf{R_t}},\mathrm{\mathbf{X}},\mathrm{\mathbf{X}})}\right),
\end{equation}
\begin{equation}
 \mathrm{\mathbf{f_{x{\rightarrow}t}}} = \mathrm{AvgPool}\left(\mathrm{CA(\mathrm{\mathbf{X}},\mathrm{\mathbf{R_t}},\mathrm{\mathbf{R_t}})}\right),
\end{equation}
where $\mathrm{AvgPool(\cdot)}$ is the average pooling over the token representations outputted by cross-attention to obtain one-vector text representation $\mathrm{\mathbf{f}}$.

\subsubsection{LLM Judgement Prediction}
Understanding the judgment hinted by the given rationale is a prerequisite for fully exploiting the information behind the rationale. To this end, we construct the LLM judgment prediction task, whose requirement is to predict the LLM judgment of the news veracity according to the given rationale. We expect this to deepen the understanding of the rationale texts.
For the textual description rationale branch, we feed its representation $\mathrm{\mathbf{R_t}}$ into the LLM judgment predictor, which is parametrized using a multi-layer perception (MLP)\footnote{For brevity, we omit the subscripts of all independently parametrized MLPs.}:
\begin{equation}
\hat{m}_t = \mathrm{sigmoid}(\mathrm{MLP}(\mathrm{\mathbf{R_t}})),
\end{equation}
\begin{equation}
L_{pt} = \mathrm{CE}(\hat{m}_t, m_t),
\end{equation}
where $m_t$ and $\hat{m}_t$ are respectively the LLM's actual judgment (extracted from the response) and its prediction. The loss $L_{pt}$ is a cross-entropy loss $\mathrm{CE}(\hat{y}, y) = -y\log \hat{y} -(1-y)\log(1-\hat{y})$. The case is similar for commonsense rationale $\mathrm{\mathbf{R_c}}$.

\subsubsection{Rationale Usefulness Evaluation}
The usefulness of rationales from different perspectives varies across different news items and improper integration may lead to performance degradation. To enable the model to adaptively select appropriate rationale, we devise a rationale usefulness evaluation process, in which we assess the contributions of different rationales and adjust their weights for subsequent veracity prediction. The process comprises two phases, \textit{i.e.}, evaluation and reweighting. For evaluation, we input the news-aware rationale vector $\mathrm{\mathbf{f_{x{\rightarrow}t}}}$ into the rationale usefulness evaluator (parameterized by an MLP) to predict its usefulness $u_t$. Following the assumption that rationales leading to correct judgments are more useful, we use the judgment correctness as the rationale usefulness labels.  

\begin{equation}
\hat{u}_t = \mathrm{sigmoid}(\mathrm{MLP}(\mathrm{\mathbf{f_{x{\rightarrow}t}}})), 
\end{equation}
\begin{equation}
L_{et} = \mathrm{CE}(\hat{u}_t, u_t).
\end{equation}
In the reweighting phase, we input vector $\mathrm{\mathbf{f_{x{\rightarrow}t}}}$ into an MLP to obtain a weight number $w_t$, which is then used to reweight the rationale-aware news vector $\mathrm{\mathbf{f_{t{\rightarrow}x}}}$. The procedure is as follows:
\begin{equation}
\mathrm{\mathbf{f_{t{\rightarrow}x}}}^{\prime} = w_t\cdot\mathrm{\mathbf{f_{t{\rightarrow}x}}}.
\end{equation}
We also use attentive pooling to transform the representation matrix $\mathrm{\mathbf{X}}$ into a vector $\mathrm{\mathbf{x}}$.

\subsection{Prediction}
Based on the outputs from the last step, we now aggregate news vector $\mathrm{\mathbf{x}}$ and rationale-aware news vector $\mathbf{f_{t{\rightarrow}x}^{\prime}}$, $\mathbf{f_{c{\rightarrow}x}^{\prime}}$ for the final judgment. For news item $x$ with label $y \in \{0,1\}$, we aggregate these vectors with different weights:
\begin{equation}
\mathrm{\mathbf{f_{cls}}}  =w_x^{cls} \cdot \mathrm{\mathbf{x}} + w_t^{cls} \cdot \mathbf{f_{t{\rightarrow}x}^{\prime}} + w_c^{cls} \cdot \mathbf{f_{c{\rightarrow}x}^{\prime}},
\end{equation}
where $w_x^{cls}$, $w_t^{cls}$ and $w_c^{cls}$ are learnable parameters ranging from 0 to 1. $\mathrm{\mathbf{f_{cls}}}$ is the fusion vector, which is then fed into the MLP classifier for final prediction of news veracity:
\begin{equation}
    L_{ce} = \mathrm{CE}(\mathrm{{MLP}}(f_{cls}), y).
\end{equation}
The total loss function is the weighted sum of the loss terms mentioned above:
\begin{equation}
    L=L_{ce}+\beta_1(L_{et}+L_{ec})+\beta_2(L_{pt}+L_{pc}),
\end{equation}
where $\beta_1$ and $\beta_2$ are hyperparameters.

\begin{table*}[t]
    \centering
    \small
    \setlength{\tabcolsep}{0.5pt}
    \begin{tabular}{@{}clcccccccc@{}}
    \toprule
      \multicolumn{2}{c}{\multirow{2}{*}[-0.3em]{\textbf{Model}}} & \multicolumn{4}{c}{\textbf{Chinese}} & \multicolumn{4}{c}{\textbf{English}} \\
      \cmidrule(lr){3-6} \cmidrule(lr){7-10}
     &   & macF1 & Acc. & F1$_\mathrm{real}$ & F1$_\mathrm{fake}$ & macF1 & Acc. & F1$_\mathrm{real}$ & F1$_\mathrm{fake}$\\
    \midrule
        \multirow{1}{*}{G1: LLM-Only} & GPT-3.5-turbo &  0.725 & 0.734 & 0.774 & 0.676 & 0.702 & 0.813 & 0.884 & 0.519  \\
        \midrule
        \multirow{4}{*}{G2: SLM-Only} & Baseline & 0.753 	& 0.754 	& 0.769 	& 0.737 	& 0.765 	& 0.862 	& 0.916 	& 0.615  \\
        &EANN$_\mathrm{T}$ & 0.754 & 0.756 & 0.773 & 0.736 & 0.763 & 0.864 & 0.918 & 0.608  \\
        &Publisher-Emo & 0.761 & 0.763 & 0.784 & 0.738 & 0.766 & 0.868 & 0.920 & 0.611  \\
        &ENDEF & 0.765 & 0.766 & 0.779 & 0.751 & 0.768 & 0.865 & 0.918 & 0.618   \\
        \midrule
        \multirow{9}{*}{G3: LLM+SLM} & Baseline + Rationale & 0.767 & 0.769 & 0.787 & 0.748 & 0.777 & 0.870 & 0.921 & 0.633  \\
        &SuperICL & 0.757 & 0.759 & 0.779 & 0.734 & 0.736 & 0.864 & 0.920 & 0.551 \\
        \cmidrule{2-10}
        &\textbf{ARG} & \textbf{0.784} & \textbf{0.786} & 0.804 & 0.764 & \textbf{0.790} & \underline{0.878} & 0.926 & 0.653 \\
        & \textit{(Relative Impr. over Baseline)} & \textit{(+4.2\%)} & \textit{(+4.3\%)} & \textit{(+4.6\%)} & \textit{(+3.8\%)} & \textit{(+3.2\%)} & \textit{(+1.8\%)} & \textit{(+1.1\%)} & \textit{(+6.3\%)} \\ 
        &\ \ w/o LLM Judgment Predictor & 0.773 & 0.774 & 0.789 & 0.756 & \underline{0.786} & \textbf{0.880} & 0.928 & 0.645   \\  
        &\ \ w/o Rationale Usefulness Evaluator & \underline{0.781} & \underline{0.783} & 0.801 & 0.761 & 0.782 & 0.873 & 0.923 & 0.641 \\    
        &\ \ w/o Predictor \& Evaluator & 0.769 & 0.770 & 0.782 & 0.756 & 0.780 & 0.874 & 0.923 & 0.637   \\   
        \cmidrule{2-10}
        & \textbf{ARG-D} & 0.771 & 0.772 & 0.785 & 0.756 & 0.778 & 0.870 & 0.921 & 0.634  \\
        & \textit{(Relative Impr. over Baseline)} & \textit{(+2.4\%)} & \textit{(+2.3\%)} & \textit{(+2.1\%)} & \textit{(+2.6\%)} & \textit{(+1.6\%)} & \textit{(+0.9\%)} & \textit{(+0.6\%)} & \textit{(+3.2\%)} \\ 
     \bottomrule
    \end{tabular}
    \caption{Performance of the ARG and its variants and the LLM-only, SLM-only, LLM+SLM methods. The best two results in macro F1 and accuracy are respectively \textbf{bolded} and \underline{underlined}. For GPT-3.5-turbo, the best results in \tablename~\ref{tab:anal_all} are reported.}
    \label{tab:main_result}
\end{table*}

\subsection{Distillation for Rationale-Free Model}
The ARG requires sending requests to the LLM for every prediction, which might not be affordable for cost-sensitive scenarios. Therefore, we attempt to build a rationale-free model, namely ARG-D, based on the trained ARG model via knowledge distillation~\cite{kd}. The basic idea is simulated and internalized the knowledge from rationales into a parametric module. As shown in \figurename~\ref{fig:arch}(d), we initialize the news encoder and classifier with the corresponding modules in the ARG and train a rationale-aware feature simulator (implemented with a multi-head transformer block) and an attention module to internalize knowledge. Besides the cross-entropy loss $L_{ce}$, we let the feature $\mathrm{\mathbf{f_{cls}^d}}$ to imitate $\mathrm{\mathbf{f_{cls}}}$ in the ARG, using the mean squared estimation loss:
\begin{equation}
    L_{kd} = \mathrm{MSE}(\mathrm{\mathbf{f_{cls}}},\mathrm{\mathbf{f_{cls}^d}}).
\end{equation}

\section{Evaluation}

\subsection{Experimental Settings}
\paragraph{Baselines}
We compare three groups of methods:

\noindent\textbf{G1 (LLM-Only)}: We list the performance of the best-performing setting on each dataset in \tablename~\ref{tab:anal_all}, \textit{i.e.}, few-shot in Chinese and few-shot CoT in English.

\noindent\textbf{G2 (SLM-Only)}\footnote{As this paper focuses on text-based news, we use the text-only variant of the original EANN following~\cite{pref-fend} and the publisher-emotion-only variant in \cite{dual-emotion}.}:
\textbf{1) Baseline:} The vanilla BERT-base model whose setting remains consistent with that in Section~\ref{sec:analysis}.
\textbf{2) EANN$_{\mathrm{T}}$~\cite{eann}:} A model that learns effective signals using auxiliary adversarial training, aiming at removing event-related features as much as possible. We used publication year as the label for the auxiliary task.
\textbf{3) Publisher-Emo~\cite{dual-emotion}:} A model that fuses a series of emotional features with textual features for fake news detection.
\textbf{4) ENDEF~\cite{endef}:} A model that removes entity bias via causal learning for better generalization on distribution-shifted fake news data. All methods in this group used the same BERT as the text encoder.

\noindent\textbf{G3 (LLM+SLM)}:
\textbf{1) Baseline+Rationale:} It concatenates features from the news encoder and rationale encoder and feeds them into an MLP for prediction.
\textbf{2) SuperICL~\cite{superICL}:} It exploits the SLM as a plug-in for the in-context learning of the LLM by injecting the prediction and the confidence for each testing sample into the prompt.

\paragraph{Implementation Details}
We use the same datasets introduced in Section~\ref{sec:analysis} and keep the setting the same in terms of the pre-trained model, learning rate, and optimization method. For the ARG-D network, the parameters of the news encoder and classifier are derived from the ARG model. A four-head transformer block is implemented in the rationale-aware feature simulator. The weight of loss functions $L_{et}, L_{pt}, L_{ec}, L_{pc}$ in the ARG and $L_{kd}$ in the ARG-D are grid searched.

\subsection{Performance Comparison and Ablation Study}
\tablename~\ref{tab:main_result} presents the performance of our proposed ARG and its variants and the compared methods. From the results, we observe that:
\textbf{1)} The ARG outperforms all other compared methods in macro F1, demonstrating its effectiveness.
\textbf{2)} The rationale-free ARG-D still outperforms all compared methods except ARG and its variants, which shows the positive impact of the distilled knowledge from ARG.
\textbf{3)} The two compared LLM+SLM methods exhibit different performance. The simple combination of features of news and rationale yields a performance improvement, showing the usefulness of our prompted rationales. SuperICL outperforms the LLM-only method 
but fails to consistently outperform the baseline SLM on the two datasets.
We speculate that this is due to the complexity of our task, where injecting prediction and confidence of an SLM does not bring sufficient information.
\textbf{4)} We evaluate three ablation experiment groups to evaluate the effectiveness of different modules in ARG network. From the result, we can see that w/o LLM Judgement Predictor or w/o Rationale Usefulness Evaluator both bring a significant decrease in ARG performance, highlighting the significance of these two structures. Besides, we found that even the weakest one among the variants of ARG still outperforms all other methods, which shows the importance of the news-rationale interaction structure we designed.

\subsection{Result Analysis}

\begin{figure}[t]
\centering
    \includegraphics[width=\linewidth]{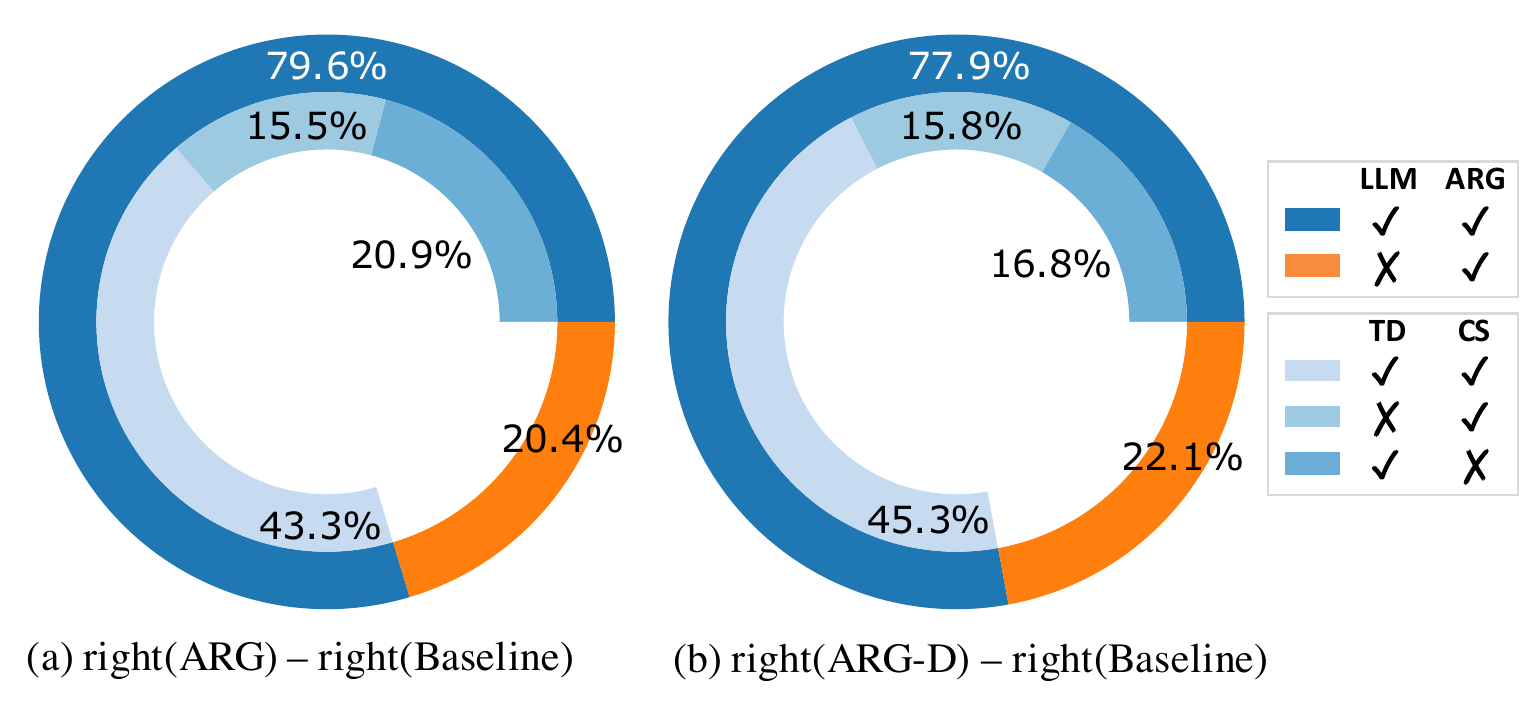}
    \caption{Statistics of additional correctly judged samples of (a) ARG and (b) ARG-D over the BERT baseline. $\mathrm{right}(\cdot)$ denotes samples correctly judged by the method $(\cdot)$. TD/CS: Textual description/commonsense perspective.}
    \label{fig:distillation}
\end{figure}

To investigate which part the additional gain of the ARG(-D) should be attributed to, we perform statistical analysis on the additional correctly judged samples of ARG(-D) compared with the vanilla BERT. From \figurename~\ref{fig:distillation}, we observe that:
\textbf{1)} The proportions of the overlapping samples between ARG(-D) and the LLM are over 77\%, indicating that the ARG(-D) can exploit (and absorb) the valuable knowledge for judgments from the LLM, even its performance is unsatisfying.
\textbf{2)} The samples correctly judged by the LLM from both two perspectives contribute the most, suggesting more diverse rationales may enhance the ARG(-D)’s training.
\textbf{3)} 20.4\% and 22.1\% of correct judgments should be attributed to the model itself. We speculate that it produces some kinds of ``new knowledge'' based on the wrong judgments of the given knowledge.

For analysis of success and failure cases and additional analysis, please refer to the appendix.

\subsection{Cost Analysis in Practice}
\begin{figure}[t]
	\centering
\includegraphics[width=0.9\linewidth]{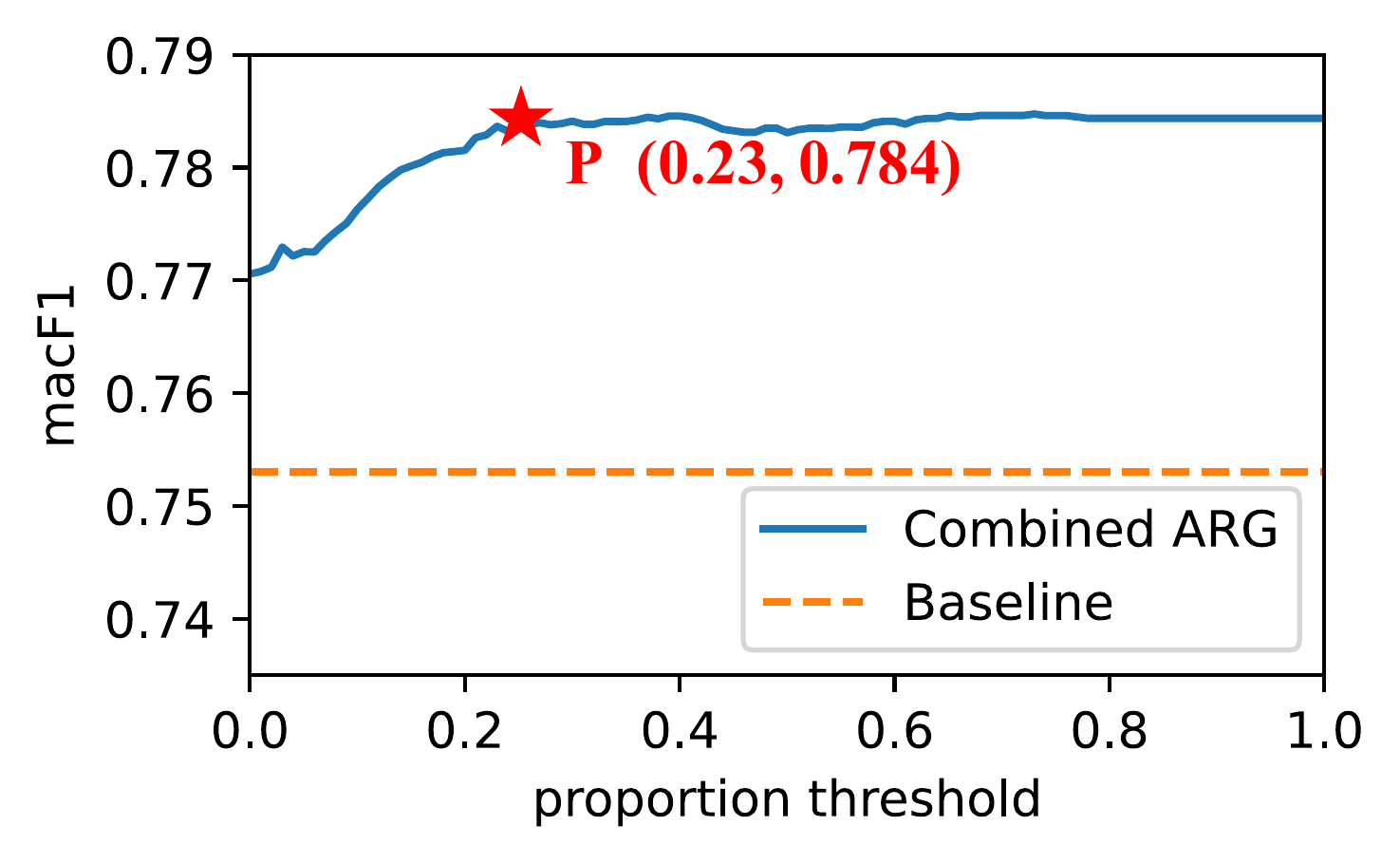}
	\caption{Performance as the shifting threshold changes.}
	\label{fig:cost}
\end{figure}

We showcase a possible model-shifting strategy to balance the performance and cost in practical systems. Inspired by~\citet{Reranker}, we simulate the situation where we use the more economic ARG-D by default but query the more powerful ARG for part of the data. As presented in \figurename~\ref{fig:cost}, by sending only 23\% of the data (according to the confidence of ARG-D) to the ARG, we could achieve 0.784 in macro F1, which is the same as the performance fully using the ARG.

\section{Related Work}
\paragraph{Fake News Detection}
\noindent Fake news detection is generally formulated as a binary classification task between real and fake news items. Research on this task could be roughly categorized into two groups: social-context-based and content-based methods. Methods in the first group aim at differentiating fake and real news during the diffusion procedure by observing the propagation patterns~\cite{network-zhou}, user feedback~\cite{dc}, and social networks~\cite{fang}. The second group focuses on finding hints based on the given content, including text~\cite{style-aaai} and images~\cite{qipeng2021} and may require extra assistance from knowledge bases~\cite{declare} and news environments~\cite{newsenv}.
Both two groups of methods obtain textual representation from pre-trained models like BERT as a convention but rarely consider its potential for fake news detection. We conducted an exploration in this paper by combining large and small LMs and obtained good improvement only using textual content.

\paragraph{LLMs for Natural Language Understanding} 
\noindent LLMs, though mostly generative models, also have powerful natural language understanding (NLU) capabilities, especially in the few-shot in-context learning scenarios~\cite{gpt3}. Recent works in this line focus on benchmarking the latest LLM in NLU. Results show that LLMs may not have comprehensive superiority compared with a well-trained small model in some types of NLU tasks~\cite{zhong2023chatgpt,if}. Our results provide empirical findings in fake news detection with only textual content as the input.

\section{Conclusion and Discussion}
We investigated if large LMs help in fake news detection and how to properly utilize their advantages for improving performance.
Results show that the large LM (GPT-3.5) underperforms the task-specific small LM (BERT), but could provide informative rationales and complement small LMs in news understanding. 
Based on these findings, we designed the ARG network to flexibly combine the respective advantages of small and large LMs and developed its rationale-free version ARG-D for cost-sensitive scenarios. Experiments showed the superiority of the ARG and ARG-D.

\paragraph{Discussion}
Our findings in fake news detection exemplify the current barrier for LLMs to be competent in applications closely related to the sophisticated real-world background.
Though having superior analyzing capability, LLMs may struggle to properly make full use of their internal capability.
This suggests that ``mining'' their potential may require novel prompting techniques and a deeper understanding of its internal mechanism.
We then identified the possibility of combining small and LLMs to earn additional improvement and provided a solution especially suitable for situations where the better-performing models have to ``select good to learn'' from worse ones.
We expect our solution to be extended to other tasks and foster more effective and cost-friendly use of LLMs in the future.

\paragraph{Limitations} We identify the following limitations: 1) We do not examine other well-known LLMs (\textit{e.g.}, Claude\footnote{\url{https://claude.ai/}} and Ernie Bot\footnote{\url{https://yiyan.baidu.com/}}) due to the API unavailability for us when conducting this research; 2) We only consider the perspectives summarized from the LLM's response and there might be other prompting perspectives based on a conceptualization framework of fake news; 3) Our best results still fall behind the oracle voting integration of multi-perspective judgments in \tablename~\ref{tab:single-view-all}, indicating that rooms still exist in our line regarding performance improvements.

\section*{Acknowledgements}
The authors would like to thank the anonymous reviewers for their insightful comments. This work is supported by the National Natural Science Foundation of China (62203425), the Zhejiang Provincial Key Research and Development Program of China (2021C01164), the Project of Chinese Academy of Sciences (E141020), the Postdoctoral Fellowship Program of CPSF (GZC20232738) and the CIPSC-SMP-Zhipu.AI Large Model Cross-Disciplinary Fund.

\bibliography{custom}
\bibliographystyle{acl_natbib}

\appendix

\section{Case Analysis}
\label{sec:cases}
Tables~\ref{tab:at_least_one_corrct} and \ref{tab:no_one_corrct} present cases in the testing set. The former includes cases where at least one of the compared methods and large language model (LLM) predictions is correct, while the latter includes those in the complementary set (\textit{i.e.,} neither of them provides correct predictions).

In \tablename~\ref{tab:at_least_one_corrct}, Case 1 shows a case in which the Baseline made a correct prediction while the LLM was wrong. The ARG stayed unaffected by the probably misleading LLM rationales and maintained the correct prediction. Cases 2 and 3 show the situation where the baseline was incorrect while the LLM could be seen partially right (\textit{i.e.}, only that used the specific perspective of the two is correct).
In these cases, our ARG selected the correct rationale based on its adaptive selection capability, resulting in the correct prediction. 

In Case 4 in \tablename~\ref{tab:no_one_corrct}, ARG successfully identified the case where both the baseline and LLM failed to provide accurate prediction, which surpasses our expectations. 
After conducting a comparative analysis, we found that within the training set, the use of phrases like "compact writing style" in textual description rationales often leads to incorrect judgments. We hypothesize that guided by the signals provided by the rationale usefulness evaluator, ARG recognized that pattern and was able to make judgments contrary to erroneous rationales and chose the correct prediction. While in Case 5, ARG was unable to invert the erroneous judgment as in Case 4. We speculate that this is due to the absence of a pattern as clear as that in Case 4. Constrained by the natural language understanding performance of the BERT-base model, ARG struggled to capture more complex logical relationships.

\section{Prompting Examples}
\label{sec:prompts}
In this section, we aim to showcase the prompting templates along with real examples for facilitating readers to know the settings directly.

In Tables~\ref{tab:prompt_res_zero} and~\ref{tab:prompt_res_few}, we illustrate prompting examples for zero-shot prompting, zero-shot CoT prompting, zero-shot CoT prompting with perspective-specific prompts, few-shot prompting, and few-shot CoT prompting from Chinese dataset. 

For brevity, we only present a news pair comprising one real news and one fake news as a demo for few-shot prompting and few-shot CoT prompting.
In practice, we utilize multiple sets of news pairs according to how many shots are required.

Due to a notable likelihood of eliciting refusals from GPT-3.5-turbo when using prompts related to fake news detection directly on the English dataset, we employ the role-playing technique. Specifically, we built appropriate contexts in the prompt to let the LLM be in a scene and ultimately decreased the refusal likelihood.~\cite{role-play-1,role-play-2}

\begingroup
\begin{table*}[htb]
    \centering
    \small
    \vspace{2.8mm}
    \begin{tabular}{p{0.96\linewidth}}
    \toprule
    
\textbf{Case 1: Both the Baseline and the ARG predicted correctly, while the LLM did incorrectly.}  \\
\vspace{-1mm}
\textbf{News:}  Wow! A robot in Russia was captured heroically saving a playful little girl from being crushed by a shelf~ This act went viral online, showing the robot's strength, amazing! \\
\textbf{LLM Rationale (Textual Description):} This message uses \underline{exaggerated, emotive language}, while providing \underline{no credible sources or evidence}. \\
\textbf{LLM Rationale (Commonsense):} Based on common knowledge and experience, robots have not yet reached a level where they can act autonomously and save human lives.
\vspace{1mm}
\\ \hdashline
\vspace{-1mm}
\textbf{Ground Truth:} \textit{\textbf{Real}} \\
\textbf{Baseline Pred.:} \textbf{\textit{Real}}\hspace{1em}
\textbf{LLM TD Pred.:} Fake\hspace{1em}
\textbf{LLM CS Pred.:} Fake\hspace{1em}
\textbf{ARG Pred.:} \textit{\textbf{Real}}
\\ \midrule

\textbf{Case 2: The Baseline predicted incorrectly, the ARG did correctly, and the LLM using textual description perspective also did correctly.}  \\
\vspace{-1mm}
\textbf{News:} \#NingboRevelation\# [Ningbo Xiangshan Fishermen Catch a Strange Fish, Dubbed the 'Panda of the Water']\#NingboFreshNews\# Around 7:00 PM on March 2nd, the ZXY02206 vessel, at the coordinates of 121°20'898" E, 29°09'898" N, caught a strange fish. Upon verification, the fish was identified as a juvenile wild Chinese sturgeon, measuring 64 centimeters in length and weighing approximately 2 kilograms. Around 10:00 AM on March 3rd, this juvenile wild Chinese sturgeon was returned to the embrace of the sea. \\
\textbf{LLM Rationale (Textual Description):} This message uses a more \underline{formal language} and provides \underline{specific details} about the time, location, and species of the fish caught, including its size. It also mentions that the information has been verified. \\
\textbf{LLM Rationale (Commonsense):} Based on common knowledge and experience, the Chinese sturgeon is a critically endangered and rare species, making it \underline{unlikely} to be commonly caught by ordinary fishermen.
\vspace{1mm}
\\ \hdashline
\vspace{-1mm}
\textbf{Ground Truth :} \textit{\textbf{Real}} \\
\textbf{Baseline Pred. :} Fake\hspace{1em}
\textbf{LLM TD Pred. :} \textbf{\textit{Real}}\hspace{1em}
\textbf{LLM CS Pred. :} Fake\hspace{1em}
\textbf{ARG Pred. :} \textit{\textbf{Real}}
\\ \midrule

\textbf{Case 3: The Baseline predicted incorrectly, the ARG did correctly, and the LLM using commonsense perspective also did correctly.} \\
\vspace{-1mm}
\textbf{News:} After recovery, the British Prime Minister commutes to work by bicycle every day, though he owns a second-hand Volkswagen. He is not allowed to use an official car for commuting, and since there's no fixed parking spot at the Prime Minister's Office, he opts for cycling. Two bodyguards follow him in a car but often fall behind due to traffic congestion. \\
\textbf{LLM Rationale (Textual Description):} The style of this message is \underline{straightforward and clear}, without excessive embellishment or exaggeration. The content described also appears to be realistic and credible. \\
\textbf{LLM Rationale (Commonsense):} Considering the nature of the British Prime Minister's job and security considerations, the claim that he commutes to work by bicycle every day is not very credible.
\vspace{1mm}
\\ \hdashline
\vspace{-1mm}
\textbf{Ground Truth :} \textit{\textbf{Fake}} \\
\textbf{Baseline Pred. :} Real\hspace{1em}
\textbf{LLM TD Pred. :} Real\hspace{1em}
\textbf{LLM CS Pred. :} \textbf{\textit{Fake}}\hspace{1em}
\textbf{ARG Pred. :} \textit{\textbf{Fake}}
\\

\bottomrule
    \end{tabular}
    \caption{
    Cases in the testing set where at least one in the Baseline and the LLM using perspective-specific prompting predicted correctly.
    TD/CS represents textual description/commonsense perspective, respectively. Patterns in rationales are \underline{underlined} and correct predictions are in \textbf{\textit{bold italics}}.
    }
    \label{tab:at_least_one_corrct}
\end{table*}
\endgroup

\begingroup
\begin{table*}[htb]
    \centering
    \small
    \vspace{2.8mm}
    \begin{tabular}{p{0.96\linewidth}}
    \toprule
    
\textbf{Case 4: Both the Baseline and the LLM using perspective-specific prompting made incorrect predictions, while the ARG did it correctly.} \\
\vspace{-1mm}
\textbf{News:} According to Russian media reports, recently, Japan entrusted US military transport planes stationed in Japan to transport 2,000 test kits donated by our country to Japan. However, during the transportation process, the US suddenly informed Japan that these test kits had been transported to the US mainland. A US Air Force official later explained that due to a mistake, these materials were wrongly transported to the US mainland. The Japanese side does not accept the explanation of the US military official! Do you think the actions of the US military were really due to a mistake? \\
\textbf{LLM Rationale (Textual Description):} The style of the message is \underline{quite compact}, with language that is \underline{concise and clear}. There are \underline{no apparent grammatical or spelling errors}, making it consistent with the style of news reporting. \\
\textbf{LLM Rationale (Commonsense):} Based on common experience and knowledge, it's \underline{not uncommon} for mistakes to occur during the transportation of goods. Therefore, it's more likely that the actions of the U.S. military were due to an error rather than intentional.
\vspace{1mm}
\\ \hdashline
\vspace{-1mm}
\textbf{Ground Truth:} \textit{\textbf{Fake}} \\
\textbf{Baseline Pred.:} Real\hspace{2em}
\textbf{LLM TD Pred.:} Real\hspace{2em}
\textbf{LLM CS Pred.:} Real\hspace{2em}
\textbf{ARG Pred.:} \textit{\textbf{Fake}}
\\ \midrule

\textbf{Case 5: The Baseline, the LLM using perspective-specific prompting, and the ARG all made incorrect predictions.} \\
\vspace{-1mm}
\textbf{News:} \#ShenzhenEpidemic\# Shenzhen's Sungang area has started entry and exit controls. Everyone should pay attention to wearing masks properly! Protect yourselves! The epidemic is not over yet, everyone should be extra careful. \\
\textbf{LLM Rationale (Textual Description):} The message utilizes a hashtag "\#ShenzhenEpidemic\#", which aligns with the common method of posting on social media. The tone is also \underline{quite formal}, serving as a reminder for everyone to take protective measures and expressing concern and attention towards the epidemic situation. \\
\textbf{LLM Rationale (Commonsense):} Given the current global situation regarding the pandemic, various regions are implementing measures to control the spread of the virus. Therefore, Shenzhen \underline{may also adopt} similar control measures. Additionally, the message mentions wearing masks and protecting oneself, which is in line with protective measures during the pandemic.
\vspace{1mm}
\\ \hdashline
\vspace{-1mm}
\textbf{Ground Truth:} \textit{\textbf{Fake}} \\
\textbf{Baseline Pred.:} Real\hspace{2em}
\textbf{LLM TD Pred.:} Real\hspace{2em}
\textbf{LLM CS Pred.:} Real\hspace{2em}
\textbf{ARG Pred.:} Real
\\

\bottomrule
    \end{tabular}
    \caption{
    Cases in the testing set where both the Baseline and the LLM using perspective-specific prompting made incorrect predictions. TD/CS represents textual description/commonsense perspective, respectively. Patterns in rationales are \underline{underlined} and correct predictions are in \textbf{\textit{bold italics}}.
    }
    \label{tab:no_one_corrct}
\end{table*}
\endgroup

\begingroup
\begin{table*}[htb]
    \centering
    \small
    \vspace{2.8mm}
    \begin{tabular}{p{0.96\linewidth}}
    \toprule
\textit{\ding{182}~Zero-Shot Prompting} \\
\vspace{-1mm}
\textbf{Q:} \textbf{Given the following message, predict its veracity. If it is more likely to be a real message, return 1; otherwise, return 0. Please refrain from providing ambiguous assessments such as undetermined:} \textit{Detailed photos of Xiang Liu's tendon surgery exposed. Stop complaints and please show sympathy and blessings!} \\
\textbf{A:} 1 \\ \midrule

\textit{\ding{183}~Zero-Shot CoT Prompting} \\
\vspace{-1mm}
\textbf{Q:} \textbf{Given the following message, predict its veracity. If it is more likely to be a real message, return 1; otherwise, return 0. Please refrain from providing ambiguous assessments such as undetermined:} \textit{Detailed photos of Xiang Liu's tendon surgery exposed. Stop complaints and please show sympathy and blessings!} \\
\textbf{A:} \textbf{Let's think step by step.} Firstly, this message mentions that Xiang Liu underwent tendon surgery, which can be verified through other channels. Secondly, the message refers to detailed photo exposure, which can also be verified by searching related reports. Lastly, the message urges people to show sympathy and offer blessings, which aligns with common practices. Therefore, it can be considered that this message is more likely to be true information, and the response is 1. \textbf{Therefore, the answer (arabic numerals) is} 1 \\ \midrule

\textit{\ding{184}~Zero-Shot CoT with Perspective Specific Prompting (Textual Description)} \\
\vspace{-1mm}
\textbf{Q:} \textbf{Given the following message, predict its veracity. If it is more likely to be a real message, return 1; otherwise, return 0. Please refrain from providing ambiguous assessments such as undetermined:} \textit{Detailed photos of Xiang Liu's tendon surgery exposed. Stop complaints and please show sympathy and blessings!} \\
\textbf{A:} \textbf{Let's think from the perspective of textual description.}      This message employs emotional language to convey sympathy and blessings for Xiang Liu, while also mentioning the detailed photo exposure of the surgery. Considering that such a message is likely to attract public attention and resonance, it is more likely to be true information. Return 1. \\ \midrule

\textit{\ding{185}~Zero-Shot CoT with Perspective Specific Prompting (Commonsense)} \\
\vspace{-1mm}
\textbf{Q:}\textbf{ Given the following message, predict its veracity. If it is more likely to be a real message, return 1; otherwise, return 0. Please refrain from providing ambiguous assessments such as undetermined:} \textit{Detailed photos of Xiang Liu's tendon surgery exposed. Stop complaints and please show sympathy and blessings!} \\
\textbf{A:} \textbf{Let's think from the perspective of commonsense.} Based on common knowledge and experience, usually, real surgery will not be exposed, as this involves patient privacy and medical institutions' confidentiality obligations. Therefore, this message is more likely to be false. Return 0. \\ 

\bottomrule
    \end{tabular}
    \caption{
    Zero-shot prompting, zero-shot CoT prompting, and zero-shot CoT with Perspective Specific prompting templates, along with the news case and the replies from the LLM (\textit{i.e.,} GPT-3.5-turbo). Template prompts are in \textbf{bold} and the target news is in \textit{italics}.
    }
    \label{tab:prompt_res_zero}
\end{table*}
\endgroup

\begingroup
\begin{table*}[htb]
\centering
\resizebox{\textwidth}{!}{
    \small
    \vspace{2.8mm}
    \begin{tabular}{p{0.96\linewidth}}
    \toprule

\textit{\ding{186}~Few-Shot Prompting} \\
\vspace{-1mm}
\textbf{Q:} Given the following message, predict its veracity. If it is more likely to be a real message, return 1; otherwise, return 0. Please refrain from providing ambiguous assessments such as undetermined:   [Russian Official Reiterates Denial: Land Transfer to China is False News] On the 10th local time, the Russian Ministry of Foreign Affairs stated that there is no territorial dispute between Russia and China. The claim that Russia transferred 4.7 square kilometers of land to China is false. After the Russia-China border was established, there have been no changes in the border demarcation. Previously, both the government of Hunchun, China, and the Russian border defense department had denied the land transfer news. http://t.cn/ruole6x \\
\textbf{A:} 1 \\
\textbf{Q:} Given the following message, predict its veracity. If it is more likely to be a real message, return 1; otherwise, return 0. Please refrain from providing ambiguous assessments such as undetermined:   Breaking News: Tianyi Li's successful exoneration! Just received news that the owner of the "rivuai160" bar in Sanlitun has been taken into custody by the police. The authorities have initiated a prostitution investigation on the bar. Tianyi Li and five others are expected to shift from being defendants to witnesses, potentially being released as soon as this weekend! \\
\textbf{A:} 0 \\ \hdashline
\vspace{-1mm}
\textbf{Q:} \textbf{Given the following message, predict its veracity. If it is more likely to be a real message, return 1; otherwise, return 0. Please refrain from providing ambiguous assessments such as undetermined:} \textit{Detailed photos of Xiang Liu's tendon surgery exposed. Stop complaints and please show sympathy and blessings!} \\
\textbf{A:} 1\\ \midrule

\textit{\ding{187}~Few-Shot CoT Prompting} \\
\vspace{-1mm}
\textbf{Q:} Given the following message, predict its veracity. If it is more likely to be a real message, return 1; otherwise, return 0. Please refrain from providing ambiguous assessments such as undetermined:   [Russian Official Reiterates Denial: Land Transfer to China is False News] On the 10th local time, the Russian Ministry of Foreign Affairs stated that there is no territorial dispute between Russia and China. The claim that Russia transferred 4.7 square kilometers of land to China is false. After the Russia-China border was established, there have been no changes in the border demarcation. Previously, both the government of Hunchun, China, and the Russian border defense department had denied the land transfer news. http://t.cn/ruole6x \\
\textbf{A:} Let's think step by step. Firstly, the source of the message is a Weibo link, and we need to consider the credibility of this link. Secondly, the content of the message pertains to a Russian denial of transferring land to China, and we need to assess if this aligns with reality. Lastly, we should check if the message aligns with reports from other reliable media sources. Based on this analysis, the message is more likely to be true information. The Russian Ministry of Foreign Affairs is an official entity, lending some credibility to their denial. Additionally, the message is consistent with the denials from the government of Hunchun, China, and the Russian border defense department. Therefore, return 1. Therefore, the answer (arabic numeral) is: 1. \\
\textbf{Q:} Given the following message, predict its veracity. If it is more likely to be a real message, return 1; otherwise, return 0. Please refrain from providing ambiguous assessments such as undetermined:   Breaking News: Tianyi Li's successful exoneration! Just received news that the owner of the "rivuai160" bar in Sanlitun has been taken into custody by the police. The authorities have initiated a prostitution investigation on the bar. Tianyi Li and five others are expected to shift from being defendants to witnesses, potentially being released as soon as this weekend! \\
\textbf{A:} Let's think step by step. Firstly, this message claims that Tianyi Li's exoneration was successful, but it doesn't specify the case in question and lacks any supporting evidence. Therefore, this part is unreliable. Secondly, the message states that the owner of the "rivuai160" bar in Sanlitun has been taken into custody by the police, and the authorities have initiated a prostitution investigation on the bar. This part could be true, as police investigations into establishments suspected of involvement in prostitution are not uncommon. However, we cannot verify the source and authenticity of this message, so it's not entirely trustworthy either. Lastly, the message suggests that Li Tianyi and five others are expected to shift from being defendants to witnesses and could be released as soon as this weekend. This part also lacks specific evidence, thus making it unreliable. In conclusion, the credibility of this message is low, and it's more likely to be untrue. Therefore, it should return 0. Hence, the answer (Arabic numeral) is: 0. \\ \hdashline
\vspace{-1mm}
\textbf{Q:} \textbf{Given the following message, predict its veracity. If it is more likely to be a real message, return 1; otherwise, return 0. Please refrain from providing ambiguous assessments such as undetermined:} \textit{Detailed photos of Xiang Liu's tendon surgery exposed. Stop complaints and please show sympathy and blessings!} \\
\textbf{A:} Firstly, this message mentions that Xiang Liu underwent tendon surgery, which can be verified through other channels. Secondly, the message refers to detailed photo exposure, which can also be verified by searching related reports. Lastly, the message urges people to show sympathy and offer blessings, which aligns with common practices. Therefore, it can be considered that this message is more likely to be true information, and the response is 1.\\ 

\bottomrule
    \end{tabular}}
    \caption{
    Few-shot prompting and few-shot CoT prompting templates, along with the news case and the replies from the LLM (\textit{i.e.,} GPT-3.5-turbo). Demos are above the dashed line. Below the dashed line, template prompts are in \textbf{bold} and the target news is in \textit{italics}.
    }
    \label{tab:prompt_res_few}
\end{table*}
\endgroup

\end{document}